\documentclass[11pt,letterpaper]{article}
\usepackage{emnlp2017}
\usepackage{times}
\usepackage{latexsym}

\usepackage{epsfig}
\usepackage{amsmath}
\usepackage{amssymb}
\usepackage{mathtools}
\usepackage{sg-macros}
\usepackage{booktabs}
\usepackage{subfig}
\emnlpfinalcopy

\def\emnlppaperid{169}

\title{Guided Open Vocabulary Image Captioning\\ with Constrained Beam Search}

\author{Peter Anderson\textsuperscript{1}, Basura Fernando\textsuperscript{1}, Mark Johnson\textsuperscript{2}, Stephen Gould\textsuperscript{1}\\
	\textsuperscript{1}The Australian National University, Canberra, Australia\\
	{\tt\small firstname.lastname@anu.edu.au}\\
	\textsuperscript{2}Macquarie University, Sydney, Australia\\
	{\tt\small mark.johnson@mq.edu.au}\\}

\date{}

\begin{document}

\maketitle

\begin{abstract} 
	Existing image captioning models do not generalize well to out-of-domain images containing novel scenes or objects. This limitation severely hinders the use of these models in real world applications dealing with images in the wild. We address this problem using a flexible approach that enables existing deep captioning architectures to take advantage of image taggers at test time, without re-training. Our method uses constrained beam search to force the inclusion of selected tag words in the output, and fixed, pretrained word embeddings to facilitate vocabulary expansion to previously unseen tag words. Using this approach we achieve state of the art results for out-of-domain captioning on MSCOCO (and improved results for in-domain captioning). Perhaps surprisingly, our results significantly outperform approaches that incorporate the same tag predictions into the learning algorithm. We also show that we can significantly improve the quality of generated ImageNet captions by leveraging ground-truth labels.
\end{abstract}

\section{Introduction}

\begin{figure}[t]
	\begin{center}
		\includegraphics[width=1\linewidth]{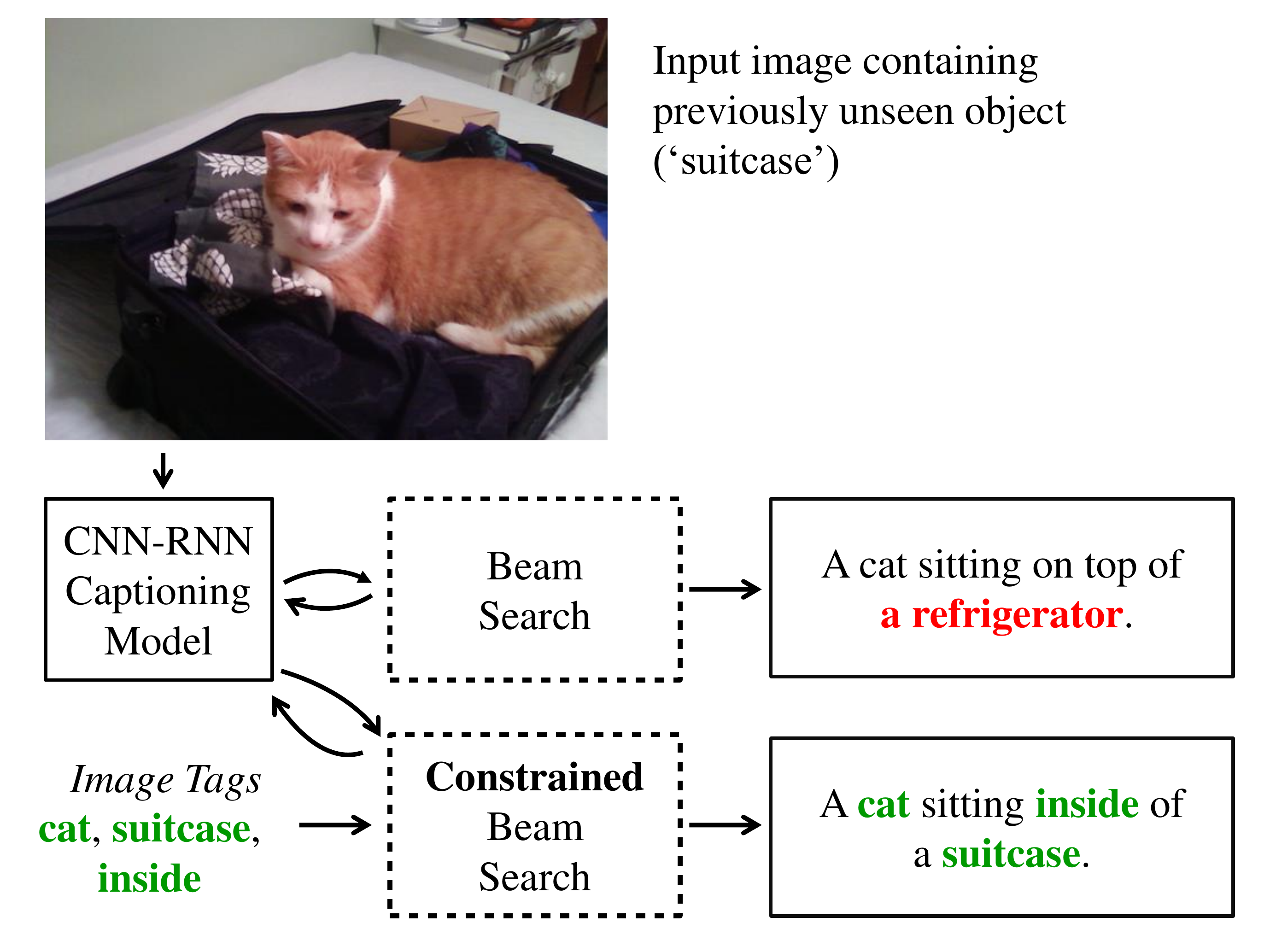}
	\end{center}
	\caption{We successfully caption images containing previously unseen objects by incorporating semantic attributes (i.e., image tags) during RNN decoding. Actual example from \secref{sec:out}.}
	\label{fig:concept}
\end{figure}

Automatic image captioning is a fundamental task that couples visual and linguistic learning. Recently, models incorporating recurrent neural networks (RNNs) have demonstrated promising results on this challenging task~\cite{Vinyals2015,Fang2015,Devlin2015b}, leveraging new benchmark datasets such as the MSCOCO dataset~\cite{Lin2014}. However, these datasets are generally only concerned with a relatively small number of objects and interactions. Unsurprisingly, models trained on these datasets do not generalize well to out-of-domain images containing novel scenes or objects~\cite{Tran2016}. This limitation severely hinders the use of these models in real world applications dealing with images in the wild. 

Although available image-caption training data is limited, many image collections are augmented with ground-truth text fragments such as semantic attributes (i.e., image tags) or object annotations. Even if these annotations do not exist, they can be generated using (potentially task specific) image taggers~\cite{chen2013fast,zhang2016fast} or object detectors~\cite{NIPS2015_5638,Krause2016}, which are easier to scale to new concepts. In this paper our goal is to incorporate text fragments such as these during caption generation, to improve the quality of resulting captions. This goal poses two key challenges. First, RNNs are generally opaque, and difficult to influence at test time. Second, text fragments may include words that are not present in the RNN vocabulary. 

As illustrated in Figure \ref{fig:concept}, we address the first challenge (guidance) by using \textit{constrained beam search} to guarantee the inclusion of selected words or phrases in the output of an RNN, while leaving the model free to determine the syntax and additional details. Constrained beam search is an approximate search algorithm capable of enforcing any constraints over resulting output sequences that can be expressed in a finite-state machine. With regard to the second challenge (vocabulary), empirically we demonstrate that an RNN can successfully generalize from similar words if both the input and output layers are fixed with pretrained word embeddings and then expanded as required.

To evaluate our approach, we use a held-out version of the MSCOCO dataset. Leveraging image tag predictions from an existing model~\cite{Hendricks2016} as constraints, we demonstrate state of the art performance for out-of-domain image captioning, while simultaneously improving the performance of the base model on in-domain data. Perhaps surprisingly, our results significantly outperform approaches that incorporate the same tag predictions into the learning algorithm~\cite{Hendricks2016,Venu2016}. Furthermore, we attempt the extremely challenging task of captioning the ImageNet classification dataset~\cite{ILSVRC15}. Human evaluations indicate that by leveraging ground truth image labels as constraints, the proportion of captions meeting or exceeding human quality increases from 11\% to 22\%. To facilitate future research we release our code and data from the project page\footnote{www.panderson.me/constrained-beam-search}.

\section{Related Work}

While various approaches to image caption generation have been considered, a large body of recent work is dedicated to neural network approaches~\cite{Donahue2015,Mao2015,Karpathy2015,Vinyals2015,Devlin2015b}. These approaches typically use a pretrained Convolutional Neural Network (CNN) image encoder, combined with a Recurrent Neural Network (RNN) decoder trained to predict the next output word, conditioned on previous words and the image. In each case the decoding process remains the same---captions are generated by searching over output sequences greedily or with beam search.

Recently, several works have proposed models intended to describe images containing objects for which no caption training data exists (out-of-domain captioning). The Deep Compositional Captioner (DCC)~\cite{Hendricks2016} uses a CNN image tagger to predict words that are relevant to an image, combined with an RNN language model to estimate probabilities over word sequences. The tagger and language models are pretrained separately, then fine-tuned jointly using the available image-caption data. 

Building on the DCC approach, the Novel Object Captioner (NOC)~\cite{Venu2016} is contemporary work with ours that also uses pretrained word embeddings in both the input and output layers of the language model. Another recent work~\cite{Tran2016} combines specialized celebrity and landmark detectors into a captioning system. 
More generally, the effectiveness of incorporating semantic attributes (i.e., image tags) into caption model training for in-domain data has been established by several works~\cite{Fang2015,CVPR16What,Elliot2015}. 

Overall, our work differs fundamentally from these approaches as we do not attempt to introduce semantic attributes, image tags or other text fragments into the learning algorithm. Instead, we incorporate text fragments during model decoding. To the best of our knowledge we are the first to consider this more loosely-coupled approach to out-of-domain image captioning, which allows the model to take advantage of information not available at training time, and avoids the need to retrain the captioning model if the source of text fragments is changed. 

More broadly, the problem of generating high probability output sequences using finite-state machinery has been previously explored in the context of poetry generation using RNNs~\cite{ghazvininejad2016generating} and machine translation using n-gram language models~\cite{allauzen2014pushdown}.

\section{Approach}

In this section we describe the constrained beam search algorithm, the base captioning model used in experiments, and our approach to expanding the model vocabulary with pretrained word embeddings.

\subsection{Constrained Beam Search}
\label{sec:cbs}

Beam search~\cite{Koehn:2010:SMT:1734086} is an approximate search algorithm that is widely used to decode output sequences from Recurrent Neural Networks (RNNs). We briefly describe the RNN decoding problem, before introducing constrained beam search, a multiple-beam search algorithm that enforces constraints in the sequence generation process.

Let $\by_t = (y_1,...,y_t)$ denote an output sequence of length $t$ containing words or other tokens from vocabulary $V$. Given an RNN modeling a probability distribution over such sequences, the RNN decoding problem is to find the output sequence with the maximum log-probability, where the log probability of any partial sequence $\by_t$ is typically given by $\sum_{j=1}^t \log p(y_{j} \mid y_1,...,y_{j-1})$. 

As it is computationally infeasible to solve this problem, beam search finds an approximate solution by maintaining a beam $B_t$ containing only the $b$ most likely partial sequences at each decoding time step $t$, where $b$ is known as the beam size. At each time step $t$, the beam $B_t$ is updated by retaining the $b$ most likely sequences in the candidate set $E_t$ generated by considering all possible next word extensions:

\begin{align}
\label{eq:bs}
E_{t} &= \big\{ (\by_{t-1},w) \mid \by_{t-1} \in B_{t-1},w \in V \big\}
\end{align}

To decode output sequences under constraints, a naive approach might impose the constraints on sequences produced at the end of beam search. However, if the constraints are non-trivial (i.e. only satisfied by relatively low probability output sequences) it is likely that an infeasibly large beam would be required in order to produce sequences that satisfy the constraints. Alternatively, imposing the constraints on partial sequences generated by \eqnref{eq:bs} is also unacceptable, as this would require that constraints be satisfied at every step during decoding---which may be impossible.

\begin{figure}[t]
	\begin{center}
		\includegraphics[trim=0cm 0cm 0cm 0cm, width=1\linewidth]{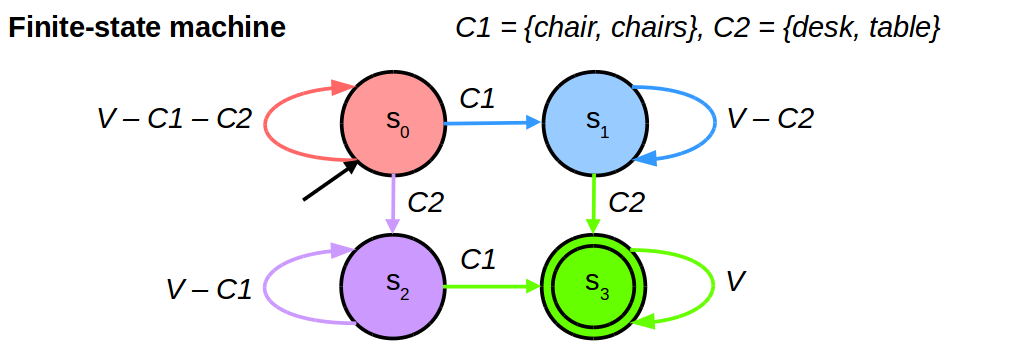}
		\includegraphics[trim=0cm 0cm 0cm 0cm, width=1\linewidth]{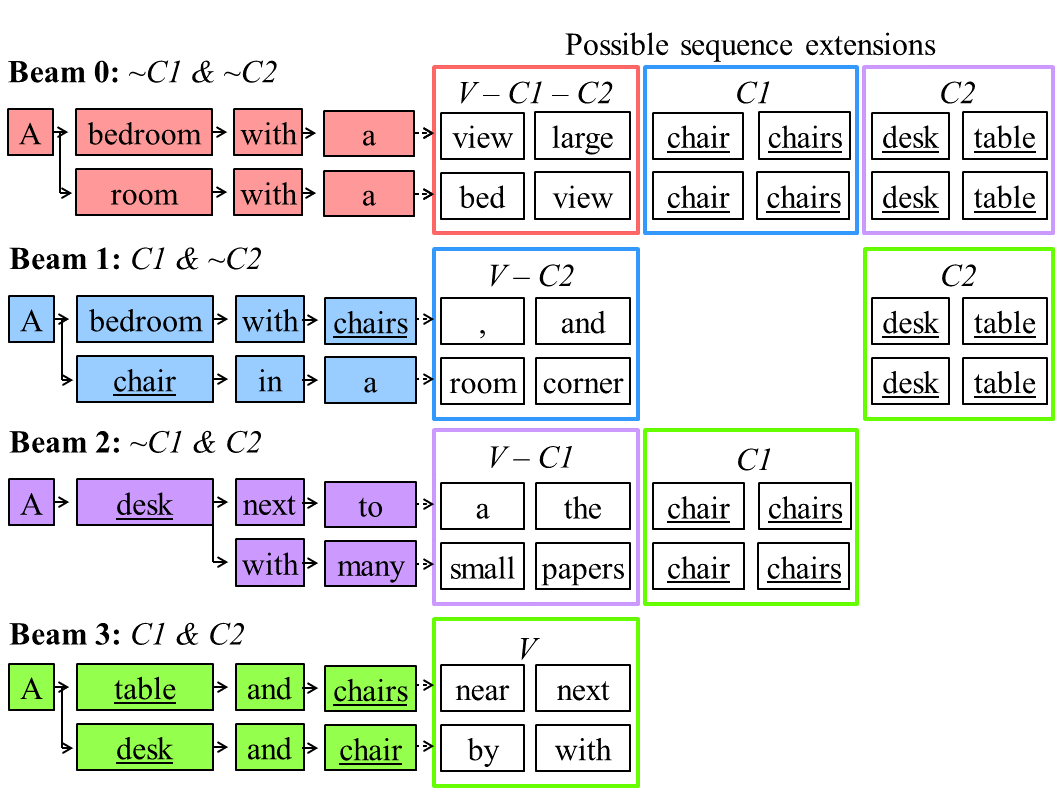}
	\end{center}
	\caption{Example of constrained beam search decoding. Each output sequence must include the words `chair' or `chairs', and `desk' or `table' from vocabulary $V$. A finite-state machine (FSM) that recognizes valid sequences is illustrated at top. Each state in the FSM corresponds to a beam in the search algorithm (bottom). FSM state transitions determine the destination beam for each possible sequence extension. Valid sequences are found in Beam 3, corresponding to FSM accepting state $s_3$.}
	\label{fig:cbs}
\end{figure}

To fix ideas, suppose that we wish to generate sequences containing at least one word from each constraint set C1 = \{`chair', `chairs'\} and C2 = \{`desk', `table'\}. Note that it is possible to \textit{recognize} sequences satisfying these constraints using the finite-state machine (FSM) illustrated in \figref{fig:cbs}, with start state $s_0$ and accepting state $s_3$. More generally, any set of constraints that can be represented with a regular expression can also be expressed as an FSM (either deterministic or non-deterministic) that recognizes sequences satisfying those constraints~\cite{Sipser:2012}. 

Since RNN output sequences are generated from left-to-right, to \textit{generate} constrained sequences, we take an FSM that recognizes sequences satisfying the required constraints, and use the following multiple-beam decoding algorithm. For each state $s \in S$ in the FSM, a corresponding search beam $B^s$ is maintained. As in beam search, each $B^s$ is a set containing at most $b$ output sequences, where $b$ is the beam size.  At each time step, each beam $B_t^s$ is updated by retaining the $b$ most likely sequences in its candidate set $E^s_{t}$ given by: 

\begin{align}
\label{eq:cbs}
\nonumber
E^s_{t} = \bigcup_{s' \in S} \big\{&(\by_{t-1},w) \mid \by_{t-1} \in B^{s'}_{t-1}, w \in V, \\ &\delta(s',w) = s  \big\}
\end{align}

\noindent
where $\delta : S \times V \mapsto S$ is the FSM state-transition function that maps states and words to states. As specified by \eqnref{eq:cbs}, the FSM state-transition function determines the appropriate candidate set for each possible extension of a partial sequence. This ensures that sequences in accepting states must satisfy all constraints as they have been recognized by the FSM during the decoding process. 

Initialization is performed by inserting an empty sequence into the beam associated with the start state $s_0$, so $B^0_0 \coloneqq \{\epsilon\}$ and $B_0^{i \neq 0} \coloneqq \emptyset$. The algorithm terminates when an accepting state contains a completed sequence (e.g., containing an end marker) with higher log probability than all incomplete sequences. In the example contained in \figref{fig:cbs}, on termination captions in Beam 0 will not contain any words from C1 or C2, captions in Beam 1 will contain a word from C1 but not C2, captions in Beam 2 will contain a word from C2 but not C1, and captions in Beam 3 will contain a word from both C1 and C2. 

\subsubsection{Implementation Details}

In our experiments we use two types of constraints. The first type of constraint consists of a conjunction of disjunctions $C = { D_1, ..., D_m }$, where each $D_i = { w_{i,1}, ..., w_{i,n_i} }$ and $w_{i,j} \in V$. Similarly to the example in \figref{fig:cbs}, a partial caption $\by_{t}$ satisfies constraint $C$ iff for each $D_i \in C$, there exists a $w_{i,j} \in D_i$ such that $w_{i,j} \in \by_{t}$. This type of constraint is used for the experiments in \secref{sec:out}, in order to allow the captioning model freedom to choose word forms. For each image tag, disjunctive sets are formed by using WordNet~\cite{wordnet} to map the tag to the set of words in $V$ that share the same lemma. 

The use of WordNet lemmas adds minimal complexity to the algorithm, as the number of FSM states, and hence the number of search beams, is not increased by adding disjunctions. Nevertheless, we note that the algorithm maintains one beam for each of the $2^m$ subsets of disjunctive constraints $D_i$. In practice $m \leq 4$ is sufficient for the captioning task, and with these values our GPU constrained beam search implementation based on Caffe~\cite{jia2014caffe} generates 40k captions for MSCOCO in well under an hour.

The second type of constraint consists of a subsequence that must appear in the generated caption. This type of constraint is necessary for the experiments in \secref{sec:imagenet}, because WordNet synsets often contain phrases containing multiple words. In this case, the number of FSM states, and the number of search beams, is linear in the length of the subsequence (the number of states is equal to number of words in a phrase $+ 1$).

\subsection{Captioning Model}

Our approach to out-of-domain image captioning could be applied to any existing CNN-RNN captioning model that can be decoding using beam search, e.g., ~\cite{Donahue2015,Mao2015,Karpathy2015,Vinyals2015,Devlin2015b}. However, for empirical evaluation we use the Long-term Recurrent Convolutional Network~\cite{Donahue2015} (LRCN) as our base model. The LRCN consists of a CNN visual feature extractor followed by two LSTM layers~\cite{Hochreiter1997}, each with 1,000 hidden units. The model is factored such that the bottom LSTM layer receives only language input, consisting of the embedded previous word. At test time the previous word is the predicted model output, but during training the ground-truth preceding word is used. The top LSTM layer receives the output of the bottom LSTM layer, as well as a per-timestep static copy of the CNN features extracted from the input image. 

The feed-forward operation and hidden state update of each LSTM layer in this model can be summarized as follows. Assuming $N$ hidden units within each LSTM layer, the $N$-dimensional input gate $i_t$, forget gate $f_t$, output gate $o_t$, and input modulation gate $g_t$ at timestep $t$ are updated as:

\begin{align}
i_t &= \sigm{W_{xi}x_t + W_{hi}h_{t-1} + b_i} \\
f_t &= \sigm{W_{xf}x_t + W_{hf}h_{t-1} + b_f} \\
o_t &= \sigm{W_{xo}x_t + W_{ho}h_{t-1} + b_o} \\
g_t &= \tanh\,(W_{xc}x_t + W_{hc}h_{t-1} + b_c)
\end{align}
\noindent
where $x_t \in \mathbb{R}^K$ is the input vector, 
$h_t \in \mathbb{R}^N$ is the LSTM output, $W$'s and $b$'s are learned weights and biases, and $\sigm{\cdot}$ and $\tanh(\cdot)$ are the sigmoid and hyperbolic tangent functions, respectively, applied element-wise. The above gates control the memory cell activation vector $c_t \in \mathbb{R}^N$ and output $h_t \in \mathbb{R}^N$ of the LSTM as follows:

\begin{align}
c_t &= f_t \odot c_{t-1} + i_t \odot g_t \\
h_t &= o_t \odot \tanh\,(c_t)
\end{align}

\noindent
where $\odot$ represents element-wise multiplication. 

Using superscripts to represent the LSTM layer index, the input vector for the bottom LSTM is an encoding of the previously generated word, given by:

\begin{align}
x_t^1 &= W_{e}\Pi_t
\end{align}

\noindent
where $W_{e}$ is a word embedding matrix, and $\Pi_t$ is a one-hot column vector identifying the input word at timestep $t$. The top LSTM input vector comprises the concatenated output of the bottom LSTM and the CNN feature descriptor of the image $I$, given by:

\begin{align}
x_t^2 &= ( h_t^1 , \text{CNN}_{\theta}(I) )
\end{align}

For the CNN component of the model, we evaluate using the 16-layer VGG~\cite{Simonyan2015} model and the 50-layer Residual Net~\cite{he2015deep}, pretrained on ILSVRC-2012~\cite{ILSVRC15} in both cases. Unlike Donahue et. al. ~\shortcite{Donahue2015}, we do not fix the CNN weights during initial training, as we find that performance improves if all training is conducted end-to-end. In training, we use only very basic data augmentation. All images are resized to 256 $\times$ 256 pixels and the model is trained on random 224 $\times$ 224 crops and horizontal flips using stochastic gradient descent (SGD) with hand-tuned learning rates. 

\subsection{Vocabulary Expansion}
\label{sec:vocab}

In the out-of-domain scenario, text fragments used as constraints may contain words that are not actually present in the captioning model's vocabulary. To tackle this issue, we leverage pretrained word embeddings, specifically the 300 dimension GloVe~\cite{pennington2014glove} embeddings trained on 42B tokens of external text corpora. 
These embeddings are introduced at both the word input and word output layers of the captioning model and fixed throughout training. Concretely, the $i$th column of the $W_{e}$ input embedding matrix is initialized with the GloVe vector associated with vocabulary word $i$. This entails reducing the dimension of the original LRCN input embedding from 1,000 to 300. The model output is then:

\begin{align}
v_t &= \tanh\,(W_{v}h_t^2 + b_{v})\\
p(y_t \mid y_{t-1},...,y_1,I) &= \text{softmax}\,(W_{e}^T v_t)
\end{align} 

\noindent
where $v_t$ represents the top LSTM output projected to 300 dimensions, $W_{e}^T$ contains GloVe embeddings as row vectors, and $p(y_t \mid y_{t-1},...,y_1,I)$ represents the normalized probability distribution over the predicted output word $y_t$ at timestep $t$, given the previous output words and the image. The model is trained with the conventional softmax cross-entropy loss function, and learns to predict $v_t$ vectors that have a high dot-product similarity with the GloVe embedding of the correct output word. 

Given these modifications --- which could be applied to other similar captioning models --- the process of expanding the model's vocabulary at test time is straightforward. To introduce an additional vocabulary word, the GloVe embedding for the new word is simply concatenated with $W_e$ as an additional column, increasing the dimension of both $\Pi_t$ and $p_t$ by one. In total there are 1.9M words in our selected GloVe embedding, which for practical purposes represents an open vocabulary. Since GloVe embeddings capture semantic and syntactic similarities~\cite{pennington2014glove}, intuitively the captioning model will generalize from similar words in order to understand how the new word can be used. 

\section{Experiments}
\label{sec:experiments}

\subsection{Microsoft COCO Dataset}

The MSCOCO 2014 captions dataset~\cite{Lin2014} contains 123,293 images, split into a 82,783 image training set and a 40,504 image validation set. Each image is labeled with five human-annotated captions. 

In our experiments we follow standard practice and perform only minimal text pre-processing, converting all sentences to lower case and tokenizing on white space. It is common practice to filter vocabulary words that occur less than five times in the training set. However, since our model does not learn word embeddings, vocabulary filtering is not necessary. Avoiding filtering increases our vocabulary from around 8,800 words to 21,689, allowing the model to potentially extract a useful training signal even from rare words and spelling mistakes (which are generally close to the correctly spelled word in embedding space). In all experiments we use a beam size of 5, and we also enforce the constraint that a single word cannot be predicted twice in a row. 

\subsection{Out-of-Domain Image Captioning}
\label{sec:out}

\begin{figure}[!h]
	\subfloat[\textbf{Base:} A woman is playing tennis on a tennis court. \textbf{Tags:} tennis, player, ball, \underline{racket}. \textbf{Base+T4:} A tennis player swinging a \underline{racket} at a ball.]{\includegraphics[width=.3\linewidth]{figures/32081.jpg}}\,\,\,
	\subfloat[\textbf{Base:} A man standing next to a yellow train. \textbf{Tags:} \underline{bus}, yellow, next, street. \textbf{Base+T4:} A man standing next to a yellow \underline{bus} on the street.]{\includegraphics[width=.3\linewidth]{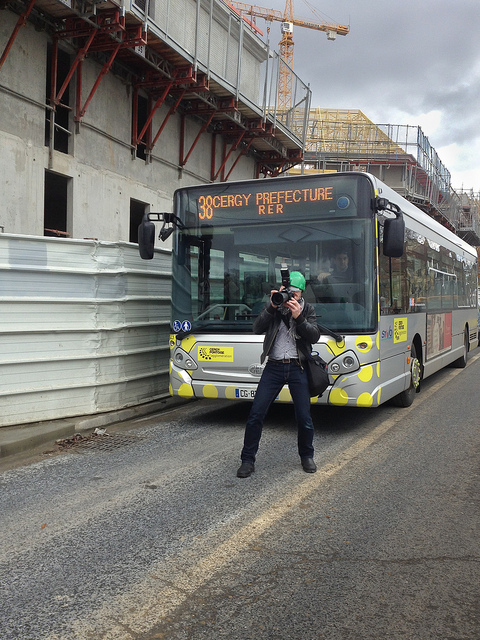}}
	\,\,\,
	\subfloat[\textbf{Base:} A close up of a cow on a dirt ground. \textbf{Tags:} \underline{zebra}, zoo, enclosure, standing. \textbf{Base+T4:} A \underline{zebra} standing in front of a zoo enclosure.]{\includegraphics[width=.3\linewidth]{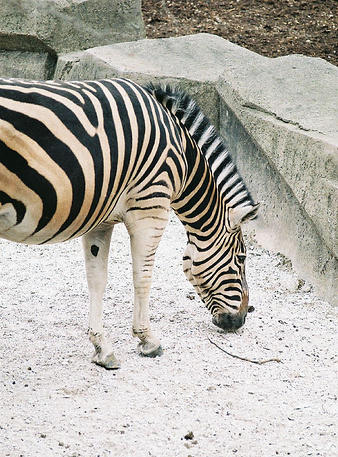}}
	\,
	
	\subfloat[\textbf{Base:} A dog is sitting in front of a tv. \textbf{Tags:} dog, head, television, cat. \textbf{Base+T4:} A dog with a cat on its head watching television.]{\includegraphics[width=.47\linewidth]{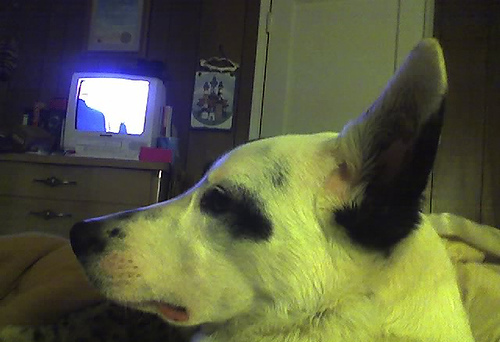}}\,\,\,
	\subfloat[\textbf{Base:} A group of people playing a game of tennis. \textbf{Tags:} pink, tennis, crowd, ball. \textbf{Base+T4:} A crowd of people standing around a pink tennis ball.]{\includegraphics[width=.47\linewidth]{figures/226959.jpg}}
	\,
	
	\caption{Examples of out-of-domain captions generated on MSCOCO using the base model (Base), and the base model constrained to include four predicted image tags (Base+T4). Words never seen in training captions are underlined. The bottom row contains some failure cases. }
	\label{fig:coco-examples}
\end{figure}

\setlength{\tabcolsep}{.52em}
\begin{table*}[t]
	\small
	\begin{center}
		\begin{tabular}{llcccccccc}
			\midrule
			&           & \multicolumn{4}{c}{Out-of-Domain Test Data} & & \multicolumn{3}{c}{In-Domain Test Data} \\
			\cmidrule{3-6}
			\cmidrule{8-10}
			Model                        & CNN       & SPICE   & METEOR   & CIDEr  & F1  & & SPICE  & METEOR  & CIDEr \\
			\midrule
			DCC~\cite{Hendricks2016}                         & VGG-16    &     13.4    &     21.0     &   59.1     &  39.8   & &   15.9     &   23.0      &   77.2      \\
			NOC~\cite{Venu2016}                       & VGG-16    &   -      &     21.4     &    -    &   49.1  & &   -     &    -     &   -     \\
			Base              & VGG-16    &     12.4    &     20.4     &    57.7    &  0   & &    17.6    &    24.9     &     93.0      \\
			Base+T1                & VGG-16    &     13.6    &   21.7       &   68.9     &  27.2  & &    17.9    &   \textbf{25.0}      &     \textbf{93.4}    \\
			Base+T2                & VGG-16    &     14.8    &     22.6     &    75.4    &  38.7   & &    \textbf{18.2}    &   \textbf{25.0}      &    92.8      \\
			Base+T3                & VGG-16    &  15.5       &    23.0      &   77.5     &  48.4   & &   \textbf{18.2}     &    24.8     &    90.4      \\
			
			Base+T4                & VGG-16    &  \textbf{15.9}       &    \textbf{23.3}      &   \textbf{77.9}     &  \textbf{54.0}   & &   18.0     &    24.5     &    86.3      \\
			
			\midrule
			Base+T3*                & VGG-16    &     18.7    &     27.1     &    119.6    &  54.5   & &    22.0    &    29.4     &    135.5       \\
			Base All Data & VGG-16    &    17.8     &     25.2     &    93.8    &   59.4  & &   17.4     &    24.5     &      91.7    \\
			\midrule
			Base                     & ResNet-50 &    12.6     &     20.5     &    56.8    &   0  & &    18.2    &    24.9     &  93.2        \\
			Base+T1               & ResNet-50 &    14.2     &     21.7     &    68.1    &  27.3   & &   18.5     &   25.2      &    \textbf{94.6}       \\
			Base+T2               & ResNet-50 &     15.3    &     22.7     &    74.7    &   38.5  & &    \textbf{18.7}    &    \textbf{25.3}     &    94.1       \\
			Base+T3              & ResNet-50 &    16.0     &     23.3     &     \textbf{77.8}   &  48.2   & &    \textbf{18.7}    &     25.2    &    92.3       \\
			Base+T4               & ResNet-50 &    \textbf{16.4}    &    \textbf{23.6}      &    77.6    &  \textbf{53.3}   & &   18.4     &   24.9      &     88.0      \\
			\midrule
			Base+T3*                & ResNet-50    &   19.2      &    27.3      &   117.9     &  54.5   & &   22.3     &   29.4      &    133.7       \\
			Base All Data & ResNet-50    &    18.6     &     26.0     &    96.9    &   60.0   & &    18.0    &    25.0     &    93.8      \\
			\midrule
		\end{tabular}
	\end{center}
	\caption{Evaluation of captions generated using constrained beam search with 1 -- 4 predicted image tags used as constraints (Base+T1 -- 4). Our approach significantly outperforms both the DCC and NOC models, despite reusing the image tag predictions of the DCC model. Importantly, performance on in-domain data is not degraded but can also improve. }
	\label{tab:results}
\end{table*}

\setlength{\tabcolsep}{.62em}
\begin{table*}[t]
	\small
	\begin{center}
		\begin{tabular}{lccccccccc}
			\midrule
			Model & bottle & bus  & couch & microwave & pizza & racket & suitcase & zebra & Avg \\ 
			\midrule
			DCC~\cite{Hendricks2016}                 & 4.6    & 29.8 & 45.9  & 28.1      & 64.6  & 52.2   & 13.2     & 79.9  & 39.8 \\
			NOC~\cite{Venu2016}               & \textbf{17.8}   & \textbf{68.8} & 25.6  & 24.7      & 69.3  & \textbf{68.1 }  & 39.9     & \textbf{89.0 } & 49.1 \\
			Base+T4 &   16.3     &   67.8   &  \textbf{48.2}  &  \textbf{ 29.7  }   &     \textbf{77.2}    &   57.1  &  \textbf{49.9 }       &    85.7   &  \textbf{54.0 }   \\ 
			\midrule
		\end{tabular}
	\end{center}
	\caption{F1 scores for mentions of objects not seen during caption training. Our approach (Base+T4) reuses the top 4 image tag predictions from the DCC model but generates higher F1 scores by interpreting tag predictions as constraints. All results based on use of the VGG-16 CNN.}
	\label{tab:fscores}
\end{table*}

To evaluate the ability of our approach to perform out-of-domain image captioning, we replicate an existing experimental design~\cite{Hendricks2016} using MSCOCO. Following this approach, all images with captions that mention one of eight selected objects (or their synonyms) are excluded from the image caption training set. This reduces the size of the caption training set from 82,783 images to 70,194 images. However, the complete caption training set is tokenized as a bag of words per image, and made available as image tag training data. As such, the selected objects are unseen in the image caption training data, but not the image tag training data. The excluded objects, selected by Hendricks~et.~al.~\shortcite{Hendricks2016} from the 80 main object categories in MSCOCO, are: `bottle', `bus', `couch', `microwave', `pizza', `racket', `suitcase' and `zebra'. 

For validation and testing on this task, we use the same splits as in prior work~\cite{Hendricks2016,Venu2016}, with half of the original MSCOCO validation set used for validation, and half for testing. We use the validation set to determine hyperparameters and for early-stopping, and report all results on the test set. For evaluation the test set is split into in-domain and out-of-domain subsets, with the out-of-domain designation given to any test image that contains a mention of an excluded object in at least one reference caption.

To evaluate generated caption quality, we use the SPICE~\cite{spice2016} metric, which has been shown to correlate well with human judgment on the MSCOCO dataset, as well as the METEOR~\cite{meteor-wmt:2014} and CIDEr~\cite{Vedantam_2015_CVPR} metrics. For consistency with previously reported results, scores on out-of-domain test data are macro-averaged across the eight excluded object classes. To improve the comparability of CIDEr scores, the inverse document frequency statistics used by this metric are determined across the entire test set, rather than within subsets. On out-of-domain test data, we also report the F1 metric for mentions of excluded objects. To calculate the F1 metric, the model is considered to have predicted condition positive if the generated caption contains at least one mention of the excluded object, and negative otherwise. The ground truth is considered to be positive for an image if the excluded object in question is mentioned in any of the reference captions, and negative otherwise.

As illustrated in \tabref{tab:results}, on the out-of-domain test data, our base model trained only with image captions (Base) receives an F1 score of 0, as it is incapable of mentioned objects that do not appear in the training captions. In terms of SPICE, METEOR and CIDEr scores, our base model performs slightly worse than the DCC model on out-of-domain data, but significantly better on in-domain data. This may suggest that the DCC model achieves improvements in out-of-domain performance at the expense of in-domain scores (in-domain scores for the NOC model were not available at the time of submission).

Results marked with `+' in \tabref{tab:results} indicate that our base model has been decoded with constraints in the form of predicted image tags. However, for the fairest comparison, and because re-using existing image taggers at test time is one of the motivations for this work, we did not train an image tagger from scratch. Instead, in results T1--4 we use the top 1--4 tag predictions respectively from the VGG-16 CNN-based image tagger used in the DCC model. This model was trained by the authors to predict 471 MSCOCO visual concepts including adjectives, verbs and nouns. Examples of generated captions, including failure cases, are presented in Figure \ref{fig:coco-examples}.  

As indicated in \tabref{tab:results}, using similar model capacity, the constrained beam search approach with predicted tags significantly outperforms prior work in terms SPICE, METEOR and CIDEr scores, across both out-of-domain and in-domain test data, utilizing varying numbers of tag predictions. Overall these results suggest that, perhaps surprisingly, it may be better to incorporate image tags into captioning models during decoding rather than during training. It also appears that, while introducing image tags improves performance on both out-of-domain and in-domain evaluations, it is beneficial to introduce more tag constraints when the test data is likely to contain previously unseen objects. This reflects the trading-off of influence between the image tags and the captioning model. For example, we noted that when using two tag constraints, 36\% of generated captions were identical to the base model, but when using four tags this proportion dropped to only 3\%. 

To establish performance upper bounds, we train the base model on the complete MSCOCO training set (Base All Data). We also evaluate captions generated using our approach combined with an `oracle' image tagger consisting of the top 3 ground-truth image tags (T3*). These were determined by selecting the 3 most frequently mentioned words in the reference captions for each test image (after eliminating stop words). The very high scores recorded for this approach may motivate the use of more powerful image taggers in future work. Finally, replacing VGG-16 with the more powerful ResNet-50~\cite{he2015deep} CNN leads to modest improvements as indicated in the lower half of \tabref{tab:results}. 

Evaluating F1 scores for object mentions (see \tabref{tab:fscores}), we note that while our approach outperforms prior work when four image tags are used, a significant increase in this score should not be expected as the underlying image tagger is the same.

\subsection{Captioning ImageNet}
\label{sec:imagenet}

\begin{figure}[!h]
	\subfloat[\textbf{Base:} A close up of a pizza on the ground. \textbf{Synset:} rock crab. \textbf{Base+Synset:} A large rock crab sitting on top of a rock.]{\includegraphics[width=.47\linewidth]{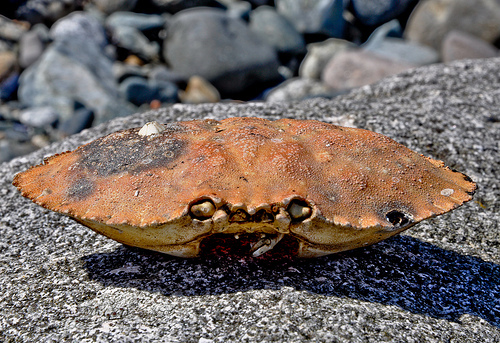}}
	\,
	\subfloat[\textbf{Base:} A close up shot of an orange. \textbf{Synset:} pool table, billiard table, snooker table. \textbf{Base+Synset:} A close up of an orange ball on a billiard table.]{\includegraphics[width=.47\linewidth]{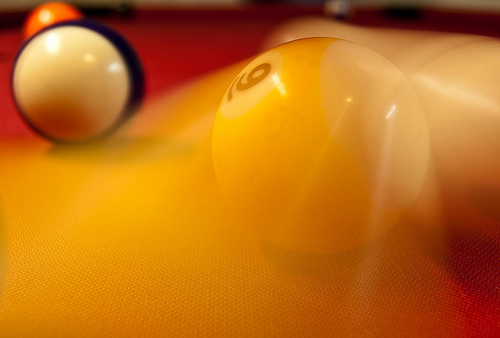}}
	\,
	\subfloat[\textbf{Base:} A herd or horses standing on a lush green field. \textbf{Synset:} \underline{rapeseed}. \textbf{Base+Synset:} A group of horses grazing in a field of  \underline{rapeseed}.]{\includegraphics[width=.47\linewidth]{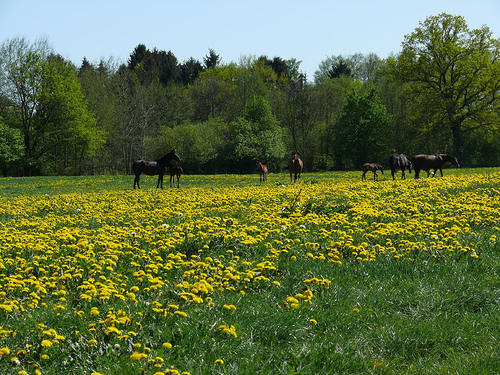}}
	\,
	\subfloat[\textbf{Base:} A black bird is standing in the grass. \textbf{Synset:} \underline{oystercatcher}, oyster catcher. \textbf{Base+Synset:} A black \underline{oystercatcher} with a red beak standing in the grass.]{\includegraphics[width=.47\linewidth]{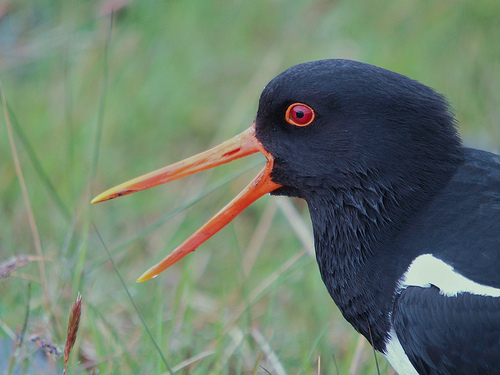}}
	\,
	\subfloat[\textbf{Base:} A man and a woman standing next to each other. \textbf{Synset:} \underline{colobus}, \underline{colobus} monkey. \textbf{Base+Synset:} Two \underline{colobus} standing next to each other near a fence.]{\includegraphics[width=.47\linewidth]{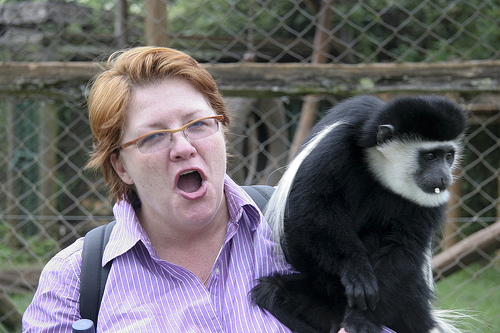}}\,\,\,
	\subfloat[\textbf{Base:} A bird standing on top of a grass covered field. \textbf{Synset:} cricket. \textbf{Base+Synset:} A bird standing on top of a cricket field.]{\includegraphics[width=.47\linewidth]{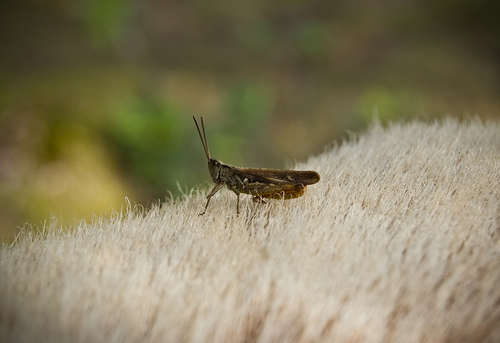}}
	\,
	\caption{Examples of ImageNet captions generated by the base model (Base), and by the base model constrained to include the ground-truth synset (Base+Synset). Words never seen in the MSCOCO / Flickr 30k caption training set are underlined. The bottom row contains some failure cases.}
	\label{fig:imagenet-examples}
\end{figure}

Consistent with our observation that many image collections contain useful annotations, and that we should seek to use this information, in this section we caption a 5,000 image subset of the ImageNet \cite{ILSVRC15} ILSVRC 2012 classification dataset for assessment. The dataset contains 1.2M images classified into 1,000 object categories, from which we randomly select five images from each category.

For this task we use the ResNet-50~\cite{he2015deep} CNN, and train the base model on a combined training set containing 155k images comprised of the MSCOCO~\cite{Chen2015} training and validation datasets, and the full Flickr 30k~\cite{young2014image} captions dataset. We use constrained beam search and vocabulary expansion to ensure that each generated caption includes a phrase from the WordNet~\cite{wordnet} synset representing the ground-truth image category. For synsets that contain multiple entries, we run constrained beam search separately for each phrase and select the predicted caption with the highest log probability overall. 

Note that even with the use of ground-truth object labels, the ImageNet captioning task remains extremely challenging as ImageNet contains a wide variety of classes, many of which are not evenly remotely represented in the available image-caption training datasets. Nevertheless, the injection of the ground-truth label frequently improves the overall structure of the caption over the base model in multiple ways. Examples of generated captions, including failure cases, are presented in Figure \ref{fig:imagenet-examples}.  

\setlength{\tabcolsep}{4.3pt}
\begin{table}[t]
	\small
	\begin{center}
		\begin{tabular}{lcccc}
			\midrule
			& Better & \begin{tabular}[c]{@{}c@{}}Equally\\ Good\end{tabular} & \begin{tabular}[c]{@{}c@{}}Equally\\ Poor\end{tabular} & Worse \\
			\midrule
			Base v. Human          & 0.05   & 0.06                                                   & 0.04                                                   & 0.86  \\	
			Base+Syn v. Human & 0.12   & 0.10                                                   & 0.05                                                   & 0.73  \\
			Base+Syn v. Base  & 0.39   & 0.06                                                   & 0.42                                                   & 0.13  \\
			\midrule
		\end{tabular}
	\end{center}
	\caption{In human evaluations our approach leveraging ground-truth synset labels (Base+Syn) improves significantly over the base model (Base) in both direct comparison and in comparison to human-generated captions. }
	\label{tab:imagenet}
\end{table}

As the ImageNet dataset contains no existing caption annotations, following the human-evaluation protocol established for the MSCOCO 2015 Captioning Challenge~\cite{Chen2015}, we used Amazon Mechanical Turk (AMT) to collect a human-generated caption for each sample image. For each of the 5,000 samples images, three human evaluators were then asked to compare the caption generated using our approach with the human-generated caption (Base+Syn v. Human). Using a smaller sample of 1,000 images, we also collected evaluations comparing our approach to the base model (Base+Syn v. Base), and comparing the base model with human-generated captions (Base v. Human). We used only US-based AMT workers, screened according to their performance on previous tasks. For both tasks, the user interface and question phrasing was identical to the MSCOCO collection process. The results of these evaluations are summarized in \tabref{tab:imagenet}.

Overall, Base+Syn captions were judged to be equally good or better than human-generated captions in 22\% of pairwise evaluations (12\% `better', 10\% `equally good'), and equally poor or worse than human-generated captions in the remaining 78\% of evaluations. Although still a long way from human performance, this is a significant improvement over the base model with only 11\% of captions judged to be equally good or better than human. For context, using the identical evaluation protocol, the top scoring model in the MSCOCO Captioning Challenge (evaluating on in-domain data) received 11\% `better', and 17\% `equally good' evaluations. 

\begin{figure}[t]
	\begin{center}
		\includegraphics[trim=0cm 1cm 0cm 0cm,width=1\linewidth]{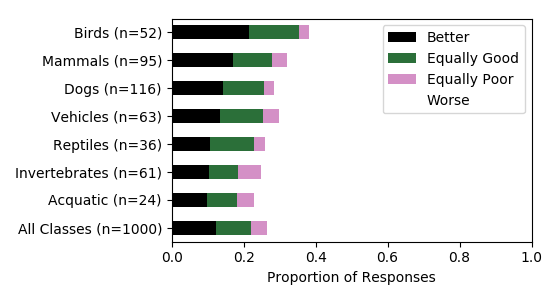}
	\end{center}
	\caption{AMT evaluations of generated (Base+Syn) ImageNet captions versus human captions, by super-category.}
	\label{fig:evals}
\end{figure}

To better understand performance across synsets, in \figref{fig:evals} we cluster some class labels into super-categories using the WordNet hierarchy, noting particularly strong performances in super-categories that have some representation in the caption training data --- such as birds, mammals and dogs. These promising results suggest that fine-grained object labels can be successfully integrated with a general purpose captioning model using our approach.

\section{Conclusion and Future Research}

We investigate \textit{constrained beam search}, an approximate search algorithm capable of enforcing any constraints over resulting output sequences that can be expressed in a finite-state machine. Applying this approach to out-of-domain image captioning on a held-out MSCOCO dataset, we leverage image tag predictions to achieve state of the art results. We also show that we can significantly improve the quality of generated ImageNet captions by using the ground-truth labels. 

In future work we hope to use more powerful image taggers, and to consider the use of constrained beam search within an expectation-maximization (EM) algorithm for learning better captioning models from weakly supervised data.

\section*{Acknowledgements}
\small
\noindent
We thank the anonymous reviewers for providing insightful comments and for helping to identify relevant prior literature. This research is supported by an Australian Government Research Training Program (RTP) Scholarship and by the Australian Research Council Centre of Excellence for Robotic Vision (project number CE140100016).

\bibliography{paper}
\bibliographystyle{emnlp_natbib}

\begin{figure*}
	\subfloat[\textbf{Base:} A couple of animals that are standing in the grass.\protect\\ 
	\textbf{Tags:} field, \underline{zebra}, grass, walking.\protect\\
	\textbf{Base+T4:} Two \underline{zebras} walking through a grass covered field.]{\includegraphics[width=.31\linewidth]{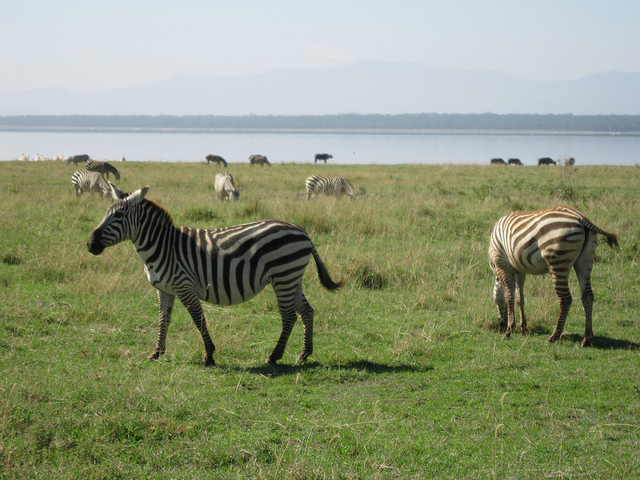}}\,\,\,
	\subfloat[\textbf{Base:} A kitchen with wooden cabinets and wooden cabinets.\protect\\ 
	\textbf{Tags:} kitchen, wooden, \underline{microwave}, oven.\protect\\
	\textbf{Base+T4:} A kitchen with wooden cabinets and a \underline{microwave} oven.]{\includegraphics[width=.31\linewidth]{figures/507958.jpg}}\,\,\,
	\subfloat[\textbf{Base:} A close up of a plate of food on a table.\protect\\ 
	\textbf{Tags:} \underline{pizza}, top, pan, cheese.\protect\\
	\textbf{Base+T4:} A \underline{pizza} pan with cheese on top of it.]{\includegraphics[width=.31\linewidth]{figures/397.jpg}}\,\,\,
	
	\subfloat[\textbf{Base:} A young boy is playing with a frisbee.\protect\\ 
	\textbf{Tags:} young, outside, wall, man.\protect\\
	\textbf{Base+T4:} A young man is outside in front of a graffiti wall.]{\includegraphics[width=.31\linewidth]{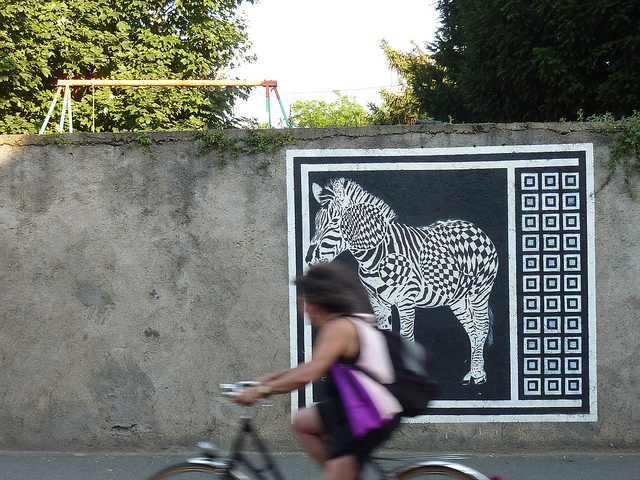}}\,\,\,
	\subfloat[\textbf{Base:} A picture of a living room with a tv.\protect\\ 
	\textbf{Tags:} room, living, \underline{couch}, window.\protect\\
	\textbf{Base+T4:} A living room with a \underline{couch} and a window.]{\includegraphics[width=.31\linewidth]{figures/33270.jpg}}\,\,\,
	\subfloat[\textbf{Base:} A man swinging a tennis ball on a tennis court.\protect\\ 
	\textbf{Tags:} \underline{racket}, tennis, court, hold.\protect\\
	\textbf{Base+T4:} A man holding a \underline{racket} on a tennis court.]{\includegraphics[width=.3\linewidth]{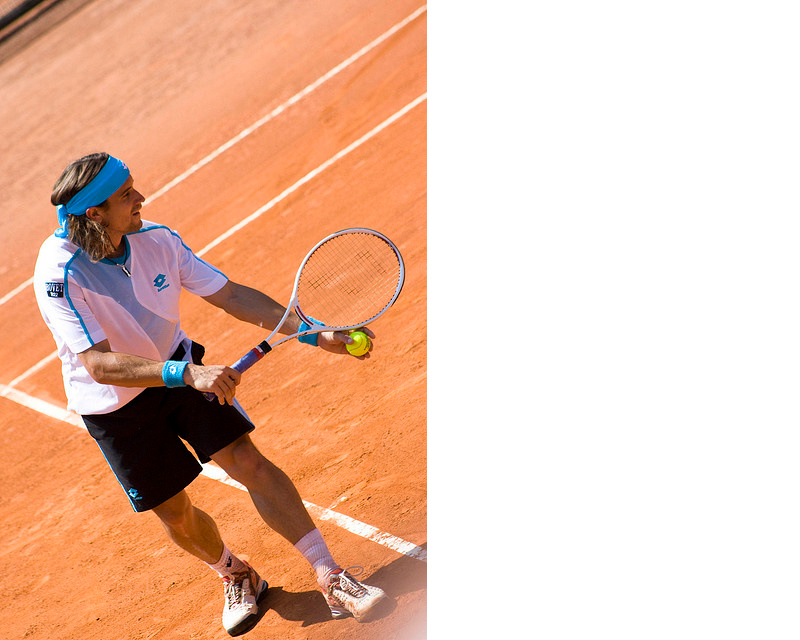}}\,\,\,
	
	\subfloat[\textbf{Base:} A close up of a microphone on a table.\protect\\ 
	\textbf{Tags:} table, \underline{bottle}, black, brown.\protect\\
	\textbf{Base+T4:} A brown and black \underline{bottle} sitting on top of a table.]{\includegraphics[width=.3\linewidth]{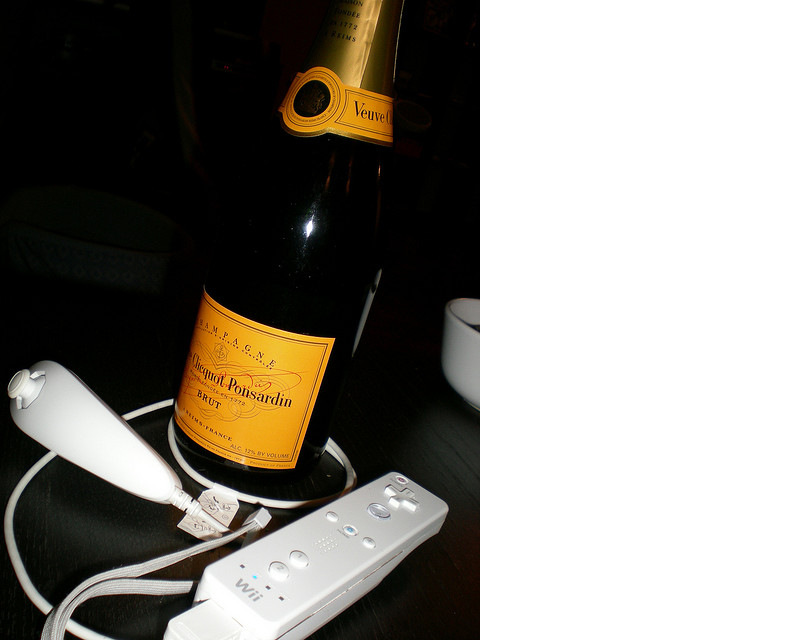}}\,\,\,
	\subfloat[\textbf{Base:} A street sign on the side of the road.\protect\\ 
	\textbf{Tags:} \underline{bus}, street, city, side.\protect\\
	\textbf{Base+T4:} A \underline{bus} is parked on the side of a city street.]{\includegraphics[width=.31\linewidth]{figures/516143.jpg}}\,\,\,
	\subfloat[\textbf{Base:} There is a toy train that is on display.\protect\\ 
	\textbf{Tags:} \underline{luggage}, cart, piled, bunch.\protect\\
	\textbf{Base+T4:} A toy cart with a bunch of \underline{luggage} piled on top of it.]{\includegraphics[width=.3\linewidth]{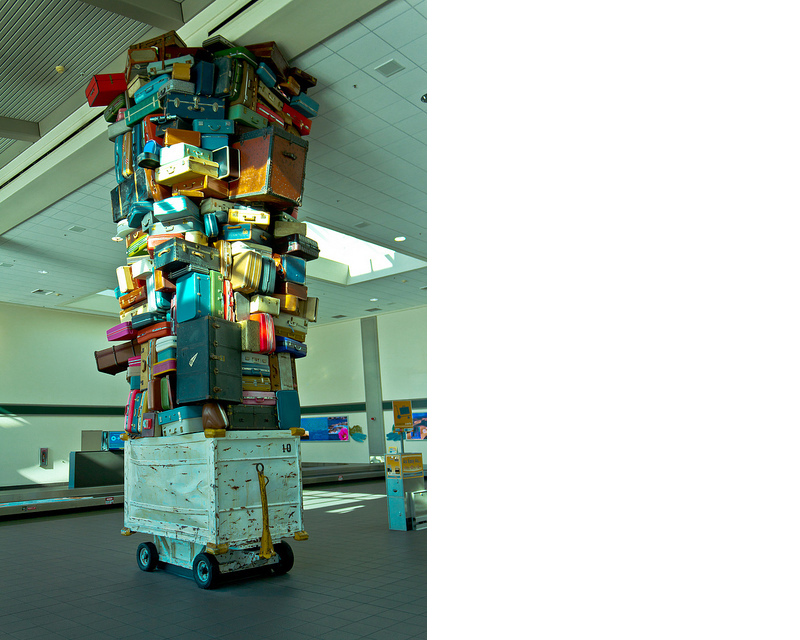}}\,\,\,
	
	\caption{Additional examples of out-of-domain captions generated on MSCOCO using the base model (Base), and the base model constrained to include four predicted image tags (Base+T4). Words never seen in training captions are underlined. }
	\label{fig:coco-examples1}
\end{figure*}

\begin{figure*}
	\subfloat[\textbf{Base:} A couple of plates of food on a table.\protect\\ 
	\textbf{Tags:} table, food, meal, dinner.\protect\\
	\textbf{Base+T4:} A meal of food on a dinner table.]{\includegraphics[width=.31\linewidth]{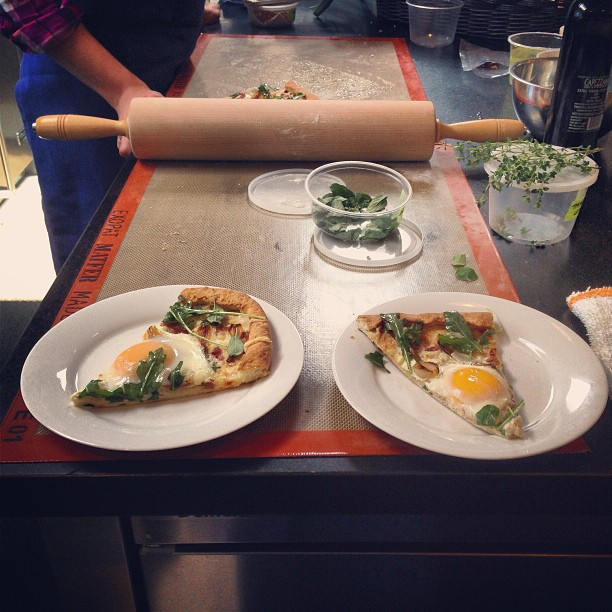}}\,\,\,
	\subfloat[\textbf{Base:} A woman is playing tennis on the beach.\protect\\ 
	\textbf{Tags:} man, dirt, player, woman.\protect\\
	\textbf{Base+T4:} A man standing on the dirt with a womens player.]{\includegraphics[width=.31\linewidth]{figures/295441.jpg}}\,\,\,
	\subfloat[\textbf{Base:} A green truck driving down a street next to a building.\protect\\ 
	\textbf{Tags:} \underline{bus}, street, green, road.\protect\\
	\textbf{Base+T4:} A green \underline{bus} truck driving down a street road.]{\includegraphics[width=.31\linewidth]{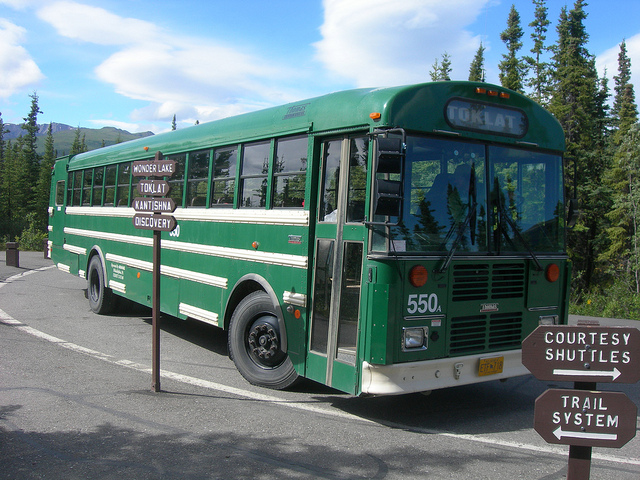}}\,\,\,
	
	\subfloat[\textbf{Base:} A man and a woman sitting on a bed.\protect\\ 
	\textbf{Tags:} bed, \underline{couch}, woman, girl.\protect\\
	\textbf{Base+T4:} A woman and a girl \underline{couch} on a bed.]{\includegraphics[width=.31\linewidth]{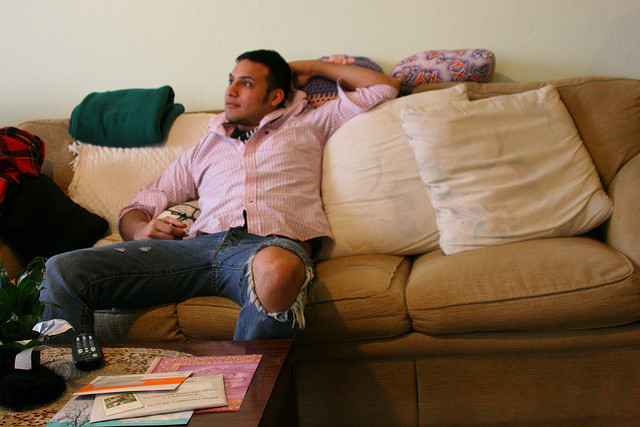}}\,\,\,
	\subfloat[\textbf{Base:} A red fire hydrant sitting in the middle of a river.\protect\\ 
	\textbf{Tags:} train, grass, field, old.\protect\\
	\textbf{Base+T4:} An old red train traveling through a grass covered field.]{\includegraphics[width=.31\linewidth]{figures/557163.jpg}}\,\,\,
	\subfloat[\textbf{Base:} A man and a woman are playing a video game.\protect\\ 
	\textbf{Tags:} hold, man, \underline{couch}, other.\protect\\
	\textbf{Base+T4:} A man holding a \underline{couch} with other people.]{\includegraphics[width=.31\linewidth]{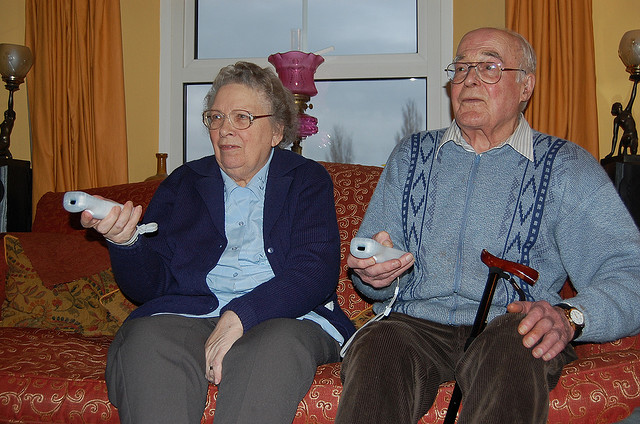}}\,\,\,
	
	\subfloat[\textbf{Base:} A man sitting at a table with a laptop.\protect\\ 
	\textbf{Tags:} table, outside, white, grass.\protect\\
	\textbf{Base+T4:} A man sitting at a table outside with white grass.]{\includegraphics[width=.31\linewidth]{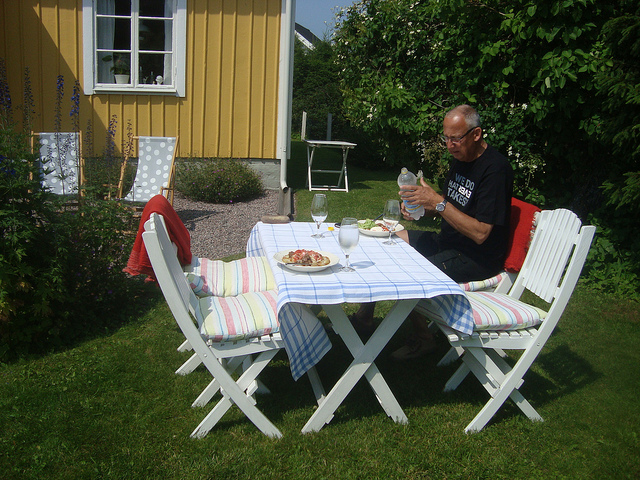}}\,\,\,
	\subfloat[\textbf{Base:} A man is playing a game of tennis.\protect\\ 
	\textbf{Tags:} \underline{racket}, tennis, man, player.\protect\\
	\textbf{Base+T4:} A man with a \underline{racket} in front of a tennis player.]{\includegraphics[width=.31\linewidth]{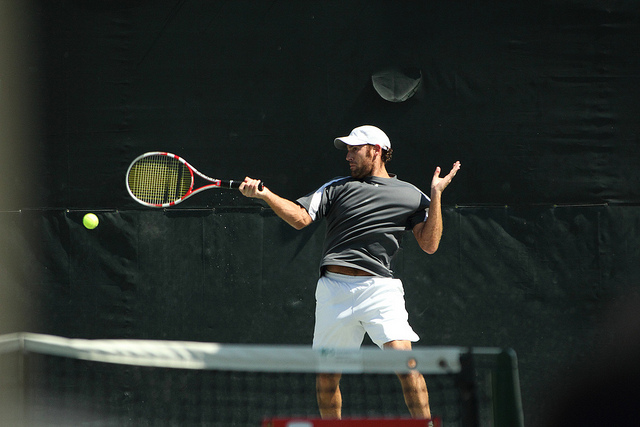}}\,\,\,
	\subfloat[\textbf{Base:} A cat that is sitting on top of a table.\protect\\ 
	\textbf{Tags:} cat, up, brown, close.\protect\\
	\textbf{Base+T4:} A close up of a cat on a brown and white cat.]{\includegraphics[width=.31\linewidth]{figures/361157.jpg}}\,\,\,
	
	\caption{Additional examples of out-of-domain MSCOCO caption failure cases, illustrating the impact of multiple image tags relating to the same object (top row), poor quality tag predictions (middle row) and other captioning errors (bottom row). Words never seen in training captions are underlined.}
	\label{fig:coco-fail}
\end{figure*}

%
%
%

\begin{figure*}
	\subfloat[\textbf{Base:} A large white boat on a body of water. \protect\\\textbf{Synset:} boathouse. \protect\\\textbf{Base + Synset:} A white and red boathouse on a lake.]{\includegraphics[width=.31\linewidth]{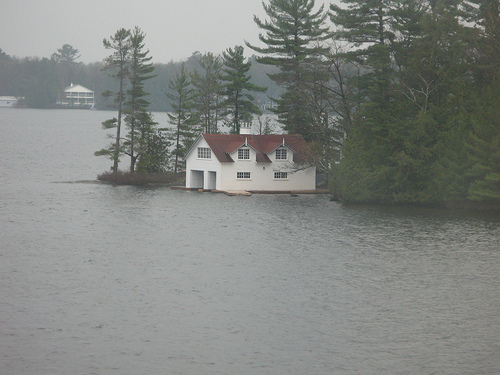}}
	\,\,
	\subfloat[\textbf{Base:} A bird perched on top of a tree branch. \protect\\\textbf{Synset:} \underline{hornbill}. \protect\\\textbf{Base + Synset:} An \underline{hornbill} bird perched on top of a tree branch.]{\includegraphics[width=.31\linewidth]{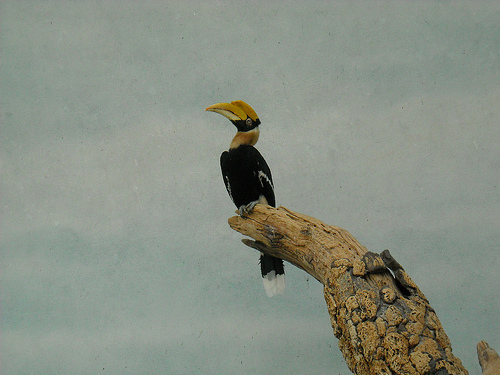}}
	\,\,
	\subfloat[\textbf{Base:} A cat sitting inside of an open refrigerator. \protect\\\textbf{Synset:} refrigerator, icebox. \protect\\\textbf{Base + Synset:} A cat sitting inside of an open refrigerator.]{\includegraphics[width=.31\linewidth]{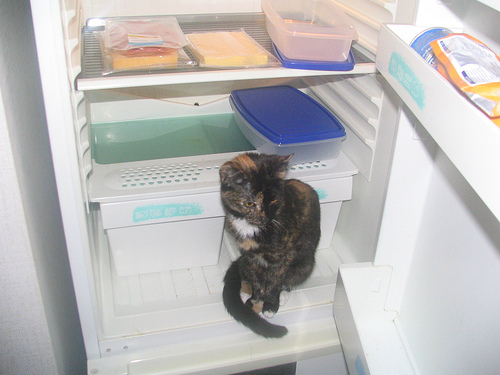}}
	\\
	\subfloat[\textbf{Base:} An old wooden suitcase sitting on a stone wall. \protect\\\textbf{Synset:} chest. \protect\\\textbf{Base + Synset:} An old wooden chest sitting next to a stone wall.]{\includegraphics[width=.31\linewidth]{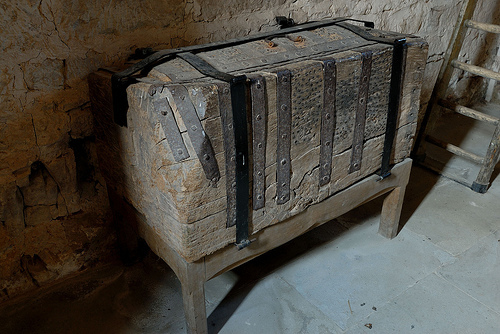}}
	\,\,
	\subfloat[\textbf{Base:} A close up of a dessert on a table. \protect\\\textbf{Synset:} \underline{trifle}. \protect\\\textbf{Base + Synset:} A close up of a \underline{trifle} cake with strawberries.]{\includegraphics[width=.31\linewidth]{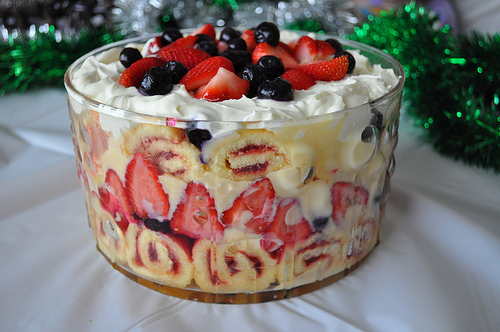}}
	\,\,
	\subfloat[\textbf{Base:} A large pile of yellow and yellow apples. \protect\\\textbf{Synset:} butternut squash. \protect\\\textbf{Base + Synset:} A pile of yellow and green butternut squash.]{\includegraphics[width=.31\linewidth]{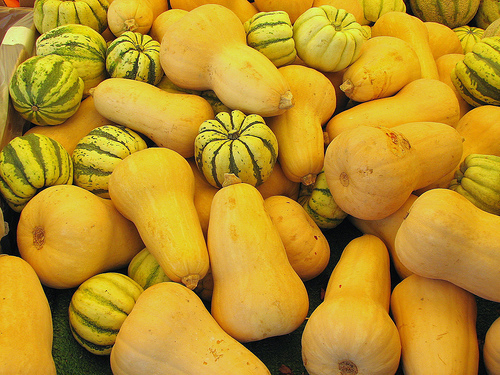}}
	\\
	\subfloat[\textbf{Base:} A black and white photo of a glass of water. \protect\\\textbf{Synset:} water bottle. \protect\\\textbf{Base + Synset:} A black and white picture of a water bottle.]{\includegraphics[width=.31\linewidth]{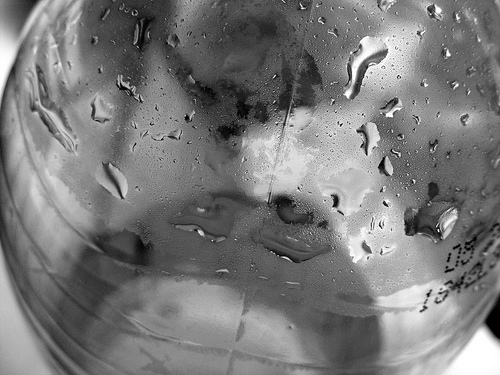}}
	\,\,
	\subfloat[\textbf{Base:} A group of animals laying on the ground. \protect\\\textbf{Synset:} \underline{Salamandra salamandra}, European fire \underline{salamander}. \protect\\\textbf{Base + Synset:} A European fire \underline{salamander} laying on the ground.]{\includegraphics[width=.31\linewidth]{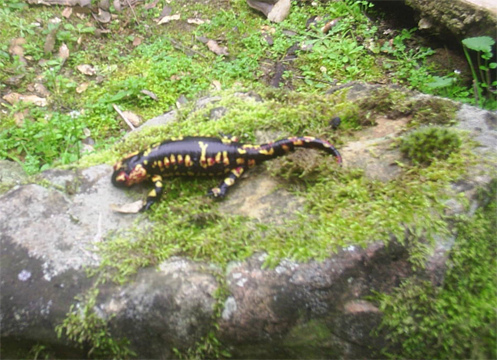}}
	\,\,
	\subfloat[\textbf{Base:} A tall tower with a clock on it. \protect\\\textbf{Synset:} triumphal arch. \protect\\\textbf{Base + Synset:} A large stone building with a triumphal arch.]{\includegraphics[width=.31\linewidth]{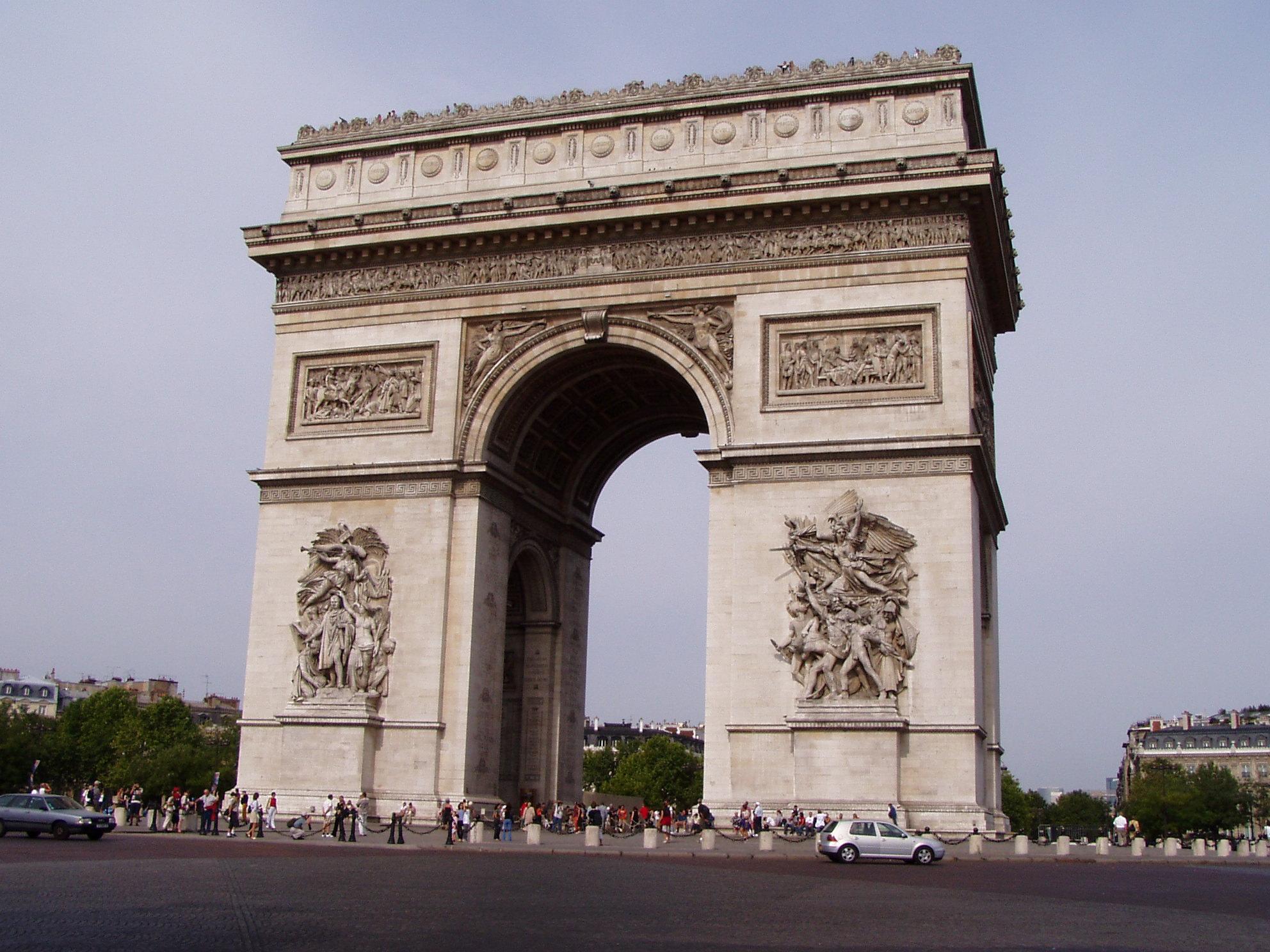}}
	\\	
	\caption{Additional examples of ImageNet captions generated by the base model (Base), and by the base model when constrained to include the ground-truth synset (Base+Synset). Occasionally the introduction of the ground-truth synset label has no effect (e.g. top right image). Words never seen in the combined MSCOCO / Flickr30k caption training set are underlined.}
	\label{fig:imagenet-more-examples}
\end{figure*}

\begin{figure*}
	\subfloat[\textbf{Base:} A teddy bear sitting on top of a table. \protect\\\textbf{Synset:} piggy bank, penny bank. \protect\\\textbf{Base + Synset:} A teddy bear sitting on a penny bank.]{\includegraphics[width=.31\linewidth]{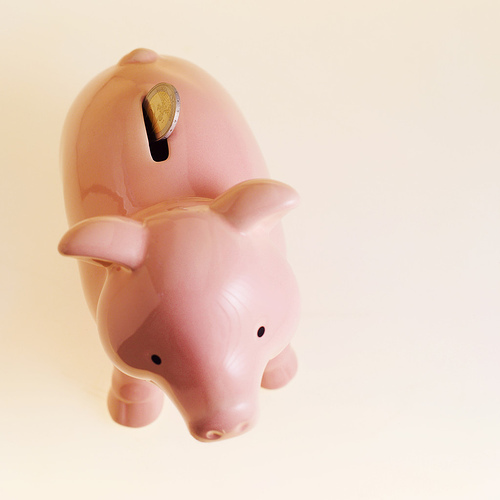}}
	\,\,
	\subfloat[\textbf{Base:} Two pictures of a dog and a dog. \protect\\\textbf{Synset:} electric ray, \underline{torpedo}. \protect\\\textbf{Base + Synset:} Two pictures of a dog and a \underline{torpedo}.]{\includegraphics[width=.31\linewidth]{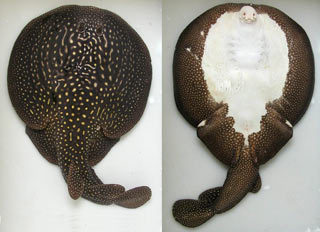}}
	\,\,
	\subfloat[\textbf{Base:} A person is feeding a cat on the ground. \protect\\\textbf{Synset:} \underline{groenendael}. \protect\\\textbf{Base + Synset:} A person holding a \underline{groenendael} and a dog.]{\includegraphics[width=.31\linewidth]{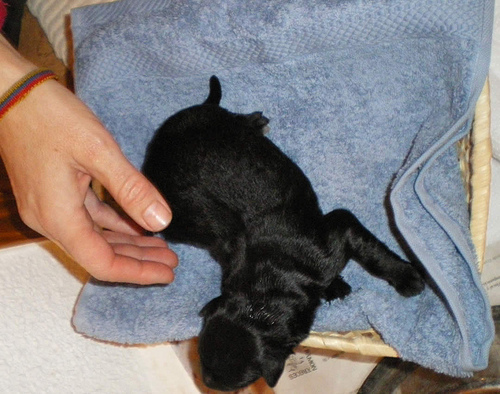}}
	\\
	\subfloat[\textbf{Base:} A large jet flying through a blue sky. \protect\\\textbf{Synset:} great white shark, white shark, \underline{man-eater}, \underline{Carcharodon carcharias}, \underline{man-eating} shark. \protect\\\textbf{Base + Synset:} A white shark flying through a blue sky.]{\includegraphics[width=.31\linewidth]{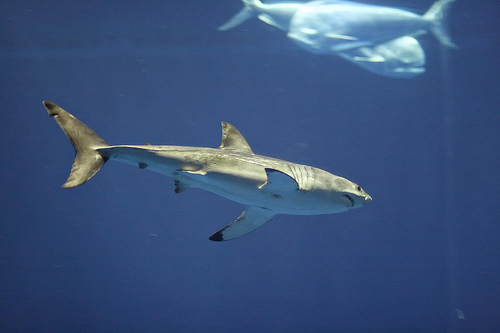}}
	\,\,
	\subfloat[\textbf{Base:} A brown dog is looking at the camera. \protect\\\textbf{Synset:} English \underline{foxhound}. \protect\\\textbf{Base + Synset:} A English \underline{foxhound} dog is looking at the camera.]{\includegraphics[width=.31\linewidth]{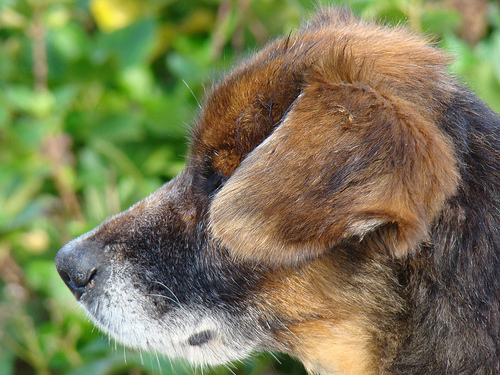}}
	\,\,
	\subfloat[\textbf{Base:} A man wearing a white shirt and tie. \protect\\\textbf{Synset:} sweatshirt. \protect\\\textbf{Base + Synset:} A man wearing a white sweatshirt and tie.]{\includegraphics[width=.31\linewidth]{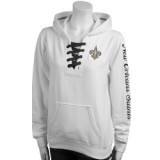}}
	\\
	\subfloat[\textbf{Base:} A man is digging a hole in the sand. \protect\\\textbf{Synset:} \underline{leatherback} turtle, \underline{leatherback}, \underline{leathery} turtle, \underline{Dermochelys coriacea}. \protect\\\textbf{Base + Synset:} A man is digging a \underline{leatherback} in sand.]{\includegraphics[width=.31\linewidth]{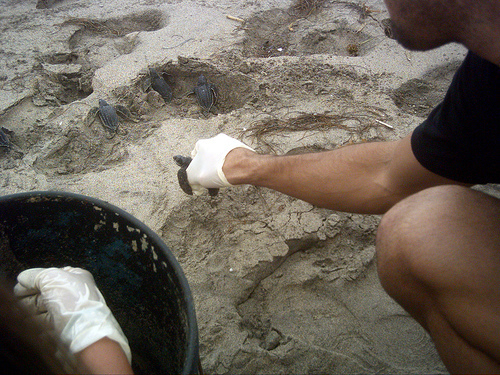}}
	\,\,
	\subfloat[\textbf{Base:} A brown and white dog laying on the grass. \protect\\\textbf{Synset:} Cardigan, Cardigan \underline{Welsh} corgi. \protect\\\textbf{Base + Synset:} A Cardigan and white dog laying on the grass.]{\includegraphics[width=.31\linewidth]{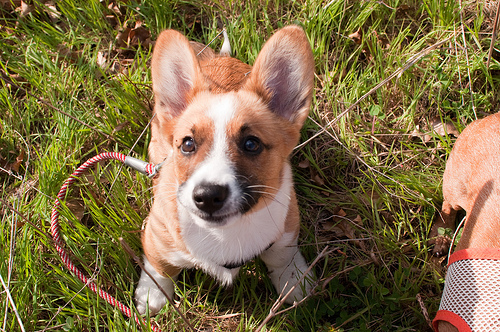}}
	\,\,
	\subfloat[\textbf{Base:} A group of stuffed animals sitting on top of each other. \protect\\\textbf{Synset:} \underline{Ambystoma mexicanum}, \underline{axolotl}, mud puppy. \protect\\\textbf{Base + Synset:} The \underline{axolotl} of two animals are on display.]{\includegraphics[width=.31\linewidth]{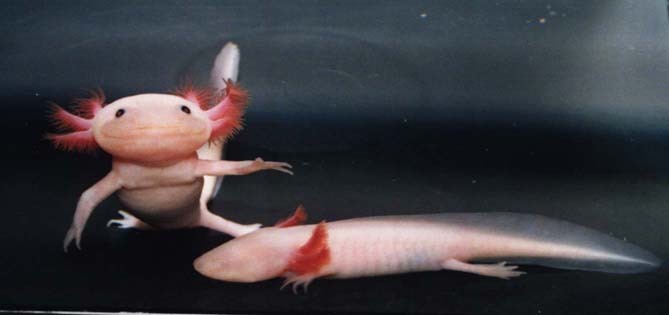}}
	\\
	\caption{Additional examples of ImageNet caption failure cases, including hallucinated objects (top), incorrect scene context (middle), and nonsensical captions (bottom). Words never seen in the combined MSCOCO / Flickr30k caption training set are underlined.}
	\label{fig:imagenet-examples1}
\end{figure*}

\end{document}